# Evaluating Large Language Models for Gait Classification Using Text-Encoded Kinematic Waveforms

Carlo Dindorf[1,*], Jonas Dully[1], Rebecca Keilhauer[2], Michael Lorenz[3], Michael Fröhlich[1]

[1]Department of Sports Science, University of Kaiserslautern-Landau (RPTU), Kaiserslautern, Germany

[2] Fraunhofer Institute for Experimental Software Engineering IESE, Kaiserslautern, Germany

[3]German Research Center of Artificial Intelligence, Kaiserslautern, Germany

**\* Correspondence:**
Carlo Dindorf
carlo.dindorf@rptu.de

## Abstract

**Background:** Machine learning (ML) enhances gait analysis but often lacks the level of interpretability desired for clinical adoption. Large Language Models (LLMs) may offer explanatory capabilities and confidence-aware outputs when applied to structured kinematic data. This study therefore evaluated whether general-purpose LLMs can classify continuous gait kinematics when represented as textual numeric sequences and how their performance compares to conventional ML approaches.

**Methods:** Lower-body kinematics were recorded from 20 participants performing seven gait patterns. A supervised KNN classifier and a class-independent One-Class SVM (OCSVM) were compared against zero-shot LLMs (GPT-5, GPT-5-mini, GPT-4.1, and o4-mini). Models were evaluated using Leave-One-Subject-Out (LOSO) cross-validation. LLMs were tested both with and without explicit reference gait statistics.

**Results:** The supervised KNN achieved the highest performance (multiclass Matthews Correlation Coefficient, MCC = 0.88). The best-performing LLM (GPT-5) with reference grounding achieved a multiclass MCC of 0.70 and a binary MCC of 0.68, outperforming the class-independent OCSVM (binary MCC = 0.60). Performance of the LLM was highly dependent on explicit reference information and self-rated confidence; when restricted to high-confidence predictions, multiclass MCC increased to 0.83 on the filtered subset. Notably, the computationally efficient o4-mini model performed comparably to larger models.

**Conclusion:** When continuous kinematic waveforms were encoded as textual numeric tokens, general-purpose LLMs, even with reference grounding, did not match supervised multiclass classifiers for precise gait classification and are better regarded as exploratory systems requiring cautious, human-guided interpretation rather than diagnostic use.

# 1. Introduction

In biomechanics, machine learning (ML) methods are increasingly employed to support clinical decision-making by providing data-driven guidance aimed at reducing diagnostic errors, shortening analysis time, enhancing reproducibility, and facilitating the interpretation of high-dimensional biomechanical datasets, thereby enabling individualized, personalized-medicine–oriented predictions beyond population-based normative references (Halilaj et al., 2018; Stetter & Stein, 2024). While ML-based classifiers have demonstrated high accuracy in biomechanical classification tasks (Balaji et al., 2020; Kwon et al., 2020; Stetter et al., 2025), their limited interpretability remains a major barrier to clinical adoption. Many state-of-the-art models operate as black boxes, offering little insight into the biomechanical features or movement patterns underlying a given classification. This lack of transparency constrains clinician trust, hinders understanding of the kinematic characteristics driving predictions, and limits the translation of model outputs into targeted treatment strategies (Horst et al., 2019; Xiang et al., 2025).

To address this limitation, the field of explainable artificial intelligence (XAI) has gained increasing attention in biomechanics (Dindorf et al., 2025; Xiang et al., 2025). Nevertheless, many existing XAI approaches generate outputs that remain difficult to interpret for clinical end users: explanations may rely on abstract, non-intuitive features, highlight overly complex interactions, or present an overwhelming amount of information that exceeds human cognitive capacity (Dindorf et al., 2020). As a result, XAI insights often remain a technical add-on rather than a practical decision-support tool. A key challenge therefore lies not only in generating explanations, but in presenting them in representations that are accessible, meaningful, and actionable for practitioners.

Recent advances in large language models (LLMs) have demonstrated their ability to transform complex and non-linguistic data representations into coherent, context-aware natural language explanations (Singh et al., 2023). Although this capability has not yet been extensively investigated for biomechanical measurement data, gait analysis data exhibit properties similar to those of high-dimensional, structured signals, indicating a promising but unexplored opportunity for LLM-based interpretation in this domain. This is supported by emerging studies demonstrating that LLMs can be used to automatically analyze biomechanical measurement data, detect abnormalities, and generate human-like reports linking kinematic findings to clinical interpretations (Zhu et al., 2025). Frameworks such as AGIR employ chain-of-thought reasoning to classify gait disorders while simultaneously providing diagnostic justifications (D. Wang et al., 2025). BiomechGPT further extends this concept by training LLMs on large-scale clinical datasets, enabling detailed responses to questions regarding movement impairments, diagnoses, and functional outcomes (Yang et al., 2025).

Previous studies indicate that OpenAI's commercial general-purpose LLMs consistently outperform many alternative models across a range of reasoning and clinical tasks and are therefore often regarded as among the most capable general-purpose LLMs currently available to end users (Szabó & Laein, 2025; Workum et al., 2025). This suggests that such models may also hold significant potential for applications in the biomechanical domain. However, these evaluations predominantly address text-based medical reasoning and decision support rather than classification tasks derived from structured biomechanical signals. To the best of the authors' knowledge, only a single pilot study has explored the use of the general-purpose OpenAI models for automated gait analysis, showing that such systems can generate clinically meaningful interpretations and are perceived as valuable decision-support tools by diverse user groups (Keilhauer et al., 2025). However, despite this initial exploratory work, systematic and quantitative evaluations of these models for gait classification remain limited.

Fundamental questions remain regarding the internal knowledge representations of state-of-the-art, general-purpose LLMs. Previous research in other domains has shown that LLM performance is highly dependent on the structure and coverage of their internal representations: different domains exhibit varying degrees of latent knowledge, and providing external grounding to compensate for missing domain knowledge can substantially improve performance (Xu et al., 2025). When such models are applied in a zero-shot setting to structured biomechanical inputs, it is currently unclear to what extent they implicitly leverage pretrained knowledge related to human movement—such as reference gait statistics of healthy gait—or whether the explicit provision of reference statistics is required to enable robust and clinically meaningful classification.

Another largely unexplored aspect concerns confidence-aware decision-making in the context of LLM-based biomechanical analysis. In conventional machine learning, confidence estimation plays a crucial role in identifying accurate predictions and flagging uncertain cases (W. He et al., 2023). Recent studies evaluating the relationship between LLMs' self-reported confidence and prediction correctness suggest that, while such models can produce explicit confidence estimates, their metacognitive accuracy varies considerably across tasks and model types (Cash et al., 2025). If high self-rated confidence aligns with higher classification accuracy, these models could be selectively integrated into clinical workflows; persistent overconfidence in incorrect predictions, however, poses a serious safety concern. Whether self-reported confidence ratings from general-purpose LLMs can meaningfully indicate prediction correctness when interpreting kinematic gait data remains unknown.

Against this background, the present study systematically investigates the use of OpenAI's general-purpose LLMs for the classification of biomechanical kinematic data. It examines to what extent general-purpose LLMs can classify biomechanical gait waveforms when these signals are represented as textual numeric sequences, compared to supervised and one-class machine learning approaches (RQ1). Furthermore, it evaluates how the provision of reference grounding with gait statistics influences LLM classification performance compared to prompts without such contextual information (RQ2). Finally, it analyzes whether LLM-generated self-rated confidence levels relate to prediction correctness and whether they can function as a practical filtering mechanism for human review (RQ3).

Using systematically induced gait variations as a controlled experimental testbed, this work aims to provide foundational insights that can guide the future integration of LLMs into biomechanical decision-support systems.

# 2. Methods

## 2.1. Workflow Overview and Evaluation Strategy

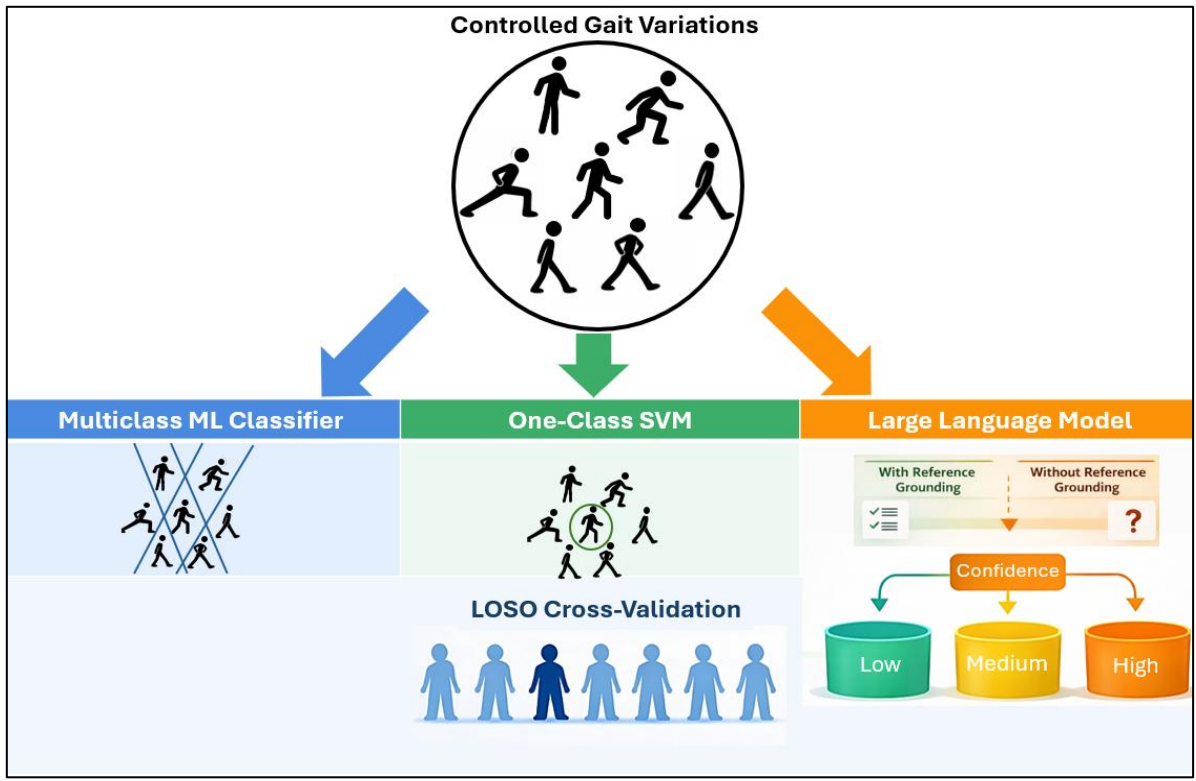

**Figure 1.** Workflow overview of the study.

Figure 1 summarizes the overall study workflow. The study is based on highly controlled, generated lower-body kinematic data representing seven gait classes. Two LLM-based approaches, one with reference grounding using information from the NORMAL gait class and one without any reference grounding, were compared to two ML approaches: (a) a supervised multiclass classifier and (b) a pathology-independent One-Class Support Vector Machine (OCSVM). In addition, both LLMs provided self-reported confidence ratings, which were used to analyze prediction accuracy as a function of confidence level.

Due to the limited number of participants, performance for all ML models was evaluated using Leave-One-Subject-Out (LOSO) cross-validation. In each iteration, all observations from one subject were held out for testing, while the remaining subjects formed the training set. This process was repeated until every subject served as the test case once. Consequently, all reported ML predictions are true out-of-fold results from models that had never seen data from the corresponding test subject. This evaluation strategy ensures strict subject-level separation and prevents data leakage.

Importantly, no actual training or fine-tuning of the LLMs was performed. The LLMs were used in an inference-only setting. Therefore, LOSO cross-validation does not apply to the LLMs in the classical sense of model training. However, a LOSO-related strategy was applied when constructing the reference information for the reference-grounded LLM. Specifically, the reference representation of the NORMAL gait class was built exclusively from data of the training subjects in each fold, explicitly excluding the NORMAL gait data of the respective test subject. Thus, while the LLM itself was not trained under a LOSO scheme, the reference data provided to it followed a subject-level exclusion principle analogous to LOSO, ensuring that no information from the test subject's NORMAL gait was included in the reference context.

## 2.2. Participants and Data Acquisition

For this study, 20 healthy adults participated (age: 23.40 ± 2.24 years; height: 173.10 ± 11.12 cm; weight: 72.6 ± 13.96 kg; 10 females, 10 males). Participants were eligible if they were free of symptoms and had no history of orthopedic or neurological disorders within the previous six months, and no acute illness at the time of testing. The sample size of the presented study is consistent with methodological norms in gait biomechanics, where many ML studies employ cohorts between 20 and 50 participants (Xiang et al., 2022) and have successfully used 20 subjects in combination with Leave-One-Subject-Out cross-validation for model evaluation (Peimankar et al., 2023). All participants were informed about the study procedures and provided written informed consent for participation and publication of the results upon arrival at the laboratory. The study was conducted in accordance with the Declaration of Helsinki and approved by the institutional ethics committee (No. 99).

Reflective markers were attached to the subjects, following the gait 2392 protocol (Delp et al., 1990). Barefoot walking was performed on a treadmill (pluto® med, h/p/cosmos sports & medical GmbH, Nussdorf-Traunstein, Germany) by all participants for five minutes at 4 km/h for familiarization. Subsequently, each participant was instructed to perform walking without and with targeted variations. The variations were adapted from (Horsak et al., 2025) and expanded to increase the number of target classes, enabling a more comprehensive evaluation of classification performance. These variations are described in Table 1 and were carried out in randomized order. Each variation was verbally instructed and corrected to meet the intended pattern. Each variation was recorded at 150 Hz using eight Qualisys cameras (6 Oqus 500+ and 2 Oqus 510+, Qualisys AB, Göteborg, Sweden). The marker trajectories were filled and filtered using a 2$^{nd}$ order low-pass Butterworth filter with a cut-off frequency of 6 Hz (Kirtley, 2006). The generic gait 2392 model was scaled (RMSE < 1cm; max error < 2cm) to the static trials of the individuals in OpenSim (v.4.5) (Delp et al., 2007). Based on the scaled models, inverse kinematics (lower body joint kinematics) were calculated.

**Table 1.** Descriptions of movement classes. For unilateral provoked movement deviations relative to normal gait, the right limb was targeted.

| Movement Class | Description |
| --- | --- |
| NORMAL | Walk with physiological gait at 4 km/h speed |
| BOUNCY | Walk in such a way that your center of mass moves strongly up and down. Do this by bending your knee deeply during the stance phase and extending it strongly at the end of the stance phase. |
| STIFF | Walk with your knees fully extended. Imagine that you have a brace on your knees that does not allow any bending. |
| LIMB_ABDUCTION | Walk normally with your left leg; your right leg rotates outward so that the toes point as far to the right as possible. Perform this movement solely from the hip, so that the hip, knee, and foot point approximately in the same direction. |
| CROUCHED | Walk with your knee and hip flexed (knee approximately 140°) and move in this squat-like position. |
| INWARD_FOOT | Turn your toes inward and maintain this position throughout the entire gait. |
| OUTWARD_FOOT | Turn your toes outward and maintain this position throughout the entire gait. |

## 2.3. Input Feature Preprocessing

Gait cycles were manually segmented, and three cycles per subject and gait class were randomly selected for analysis. The extracted kinematic features comprised pelvis tilt, obliquity, and rotation, as well as hip flexion, adduction, and rotation, knee flexion, and ankle dorsiflexion. Except for pelvis kinematics, all variables were analyzed bilaterally.

Rather than deriving summary descriptors such as range of motion or peak values, the full kinematic waveforms were used as model input (Keilhauer et al., 2025). This decision preserves the temporal structure of gait and retains phase-specific information that is essential for characterizing movement patterns. It also avoids subjective preselection of potentially informative metrics, which could introduce investigator bias and discard relevant information prior to modeling. By providing the complete kinematic waveforms, the approach allows assessment of whether the LLM can autonomously identify relevant movement characteristics during its reasoning process, without being constrained by predefined feature engineering.

Each gait cycle was time-normalized to 100% of the gait cycle. Subsequently, the waveform was resampled at eleven equally spaced time points (0%, 10%, 20%, ..., 100%) using cubic spline interpolation. This level of temporal downsampling was established through pilot analyses conducted prior to the present study, which demonstrated that the resulting representation retains the essential kinematic features required to achieve high classification accuracy across the evaluated ML models. Given that 13 kinematic features were extracted per time point, this resulted in a total feature vector size of 143 input dimensions per gait cycle.

All time-normalized features were organized in a wide-format representation in which each feature–timepoint combination constituted one input dimension. Identical input features were used for both ML and LLM classifiers; however, for the LLM, the data were provided in a structured textual format as described in detail below.

## 2.4. Supervised Multiclass Classification

A supervised multiclass classifier was trained using samples from all gait classes. Because this model had access to the complete spectrum of movement patterns during training, it captured the full discriminative structure of the dataset.

A K-Nearest Neighbors (KNN) classifier was employed to predict the multiclass target due to its low model complexity, which reduces the risk of overfitting (Peterson, 2009) and because of its demonstrated effectiveness in gait classification tasks (Eltanani et al., 2021; Mezghani et al., 2008). Furthermore, the highly controlled and simulated recording conditions were expected to simplify class discrimination, reducing the need for more complex models. In contrast to more sophisticated approaches, KNN relies directly on similarity in the feature space and thus provides a transparent reference for classification performance based purely on kinematic proximity.

Within each LOSO iteration, the KNN model was trained on data from all training subjects and evaluated on the held-out subject. Predictions from all iterations were aggregated to obtain multiclass predictions for the entire dataset. The model was implemented using scikit-learn (Buitinck et al., 2013) with default parameter settings.

## 2.5. One-Class Support Vector Machine

To enable a fair comparison with the LLM reference-grounding scenario, a second machine learning approach was designed that had access only to data from the NORMAL gait class. This setup reflects the clinical reality in which not all pathological gait patterns are known or available during model development. Following prior work on pathology-independent gait assessment (Dindorf et al., 2021; Teufl et al., 2021), a one-class classification strategy was implemented using a One-Class Support Vector Machine (OCSVM).

Accordingly, the original multiclass labels were projected into a binary decision space (NORMAL vs. NOT-NORMAL) by collapsing all non-NORMAL gait classes into a single NOT-NORMAL category. Within each LOSO iteration, the OCSVM was trained on NORMAL samples from the training subjects. As previous studies have emphasized the importance of careful hyperparameter tuning for OCSVMs (S. Wang et al., 2018), the kernel scale parameter ($\gamma$) and the expected outlier fraction ($\nu$) were optimized using nested cross-validation within the training data. Specifically, while model fitting in each inner fold was performed exclusively on NORMAL samples, the corresponding validation splits included both NORMAL and NOT-NORMAL samples from the training subjects to enable performance-based hyperparameter selection. A stratified three-fold inner cross-validation was employed, and performance was optimized using the Matthews Correlation Coefficient (MCC). After hyperparameter tuning, the OCSVM was retrained on all NORMAL samples from the training subjects and subsequently evaluated on the left-out subject.

## 2.6. LLM-Based Classification

### 2.6.1. Model Selection and General Procedure

LLMs were evaluated as zero-shot biomechanical classifiers to investigate whether general-purpose reasoning models can interpret structured kinematic waveforms without task-specific training. Two generations of large language models (GPT-5 and GPT-4.1) were evaluated to provide a comparative performance reference. Additionally, their computationally efficient counterparts (GPT-5-mini and o4-mini) were included to examine feasibility under realistic deployment constraints. All models were accessed via their OpenAI API aliases (gpt-5, gpt-4.1, gpt-5-mini, o4-mini) on 2026-02-02, which correspond to the latest available model snapshots at the time of evaluation. This selection allowed comparison between high-capacity reasoning models and smaller, more cost-efficient variants to determine whether complex biomechanical interpretation requires large model capacity or can be achieved with lighter architectures.

The LLM was accessed via a stateless API without fine-tuning or memory across samples. All models were used with their default parameter settings to ensure reproducibility and to avoid hyperparameter optimization that could introduce bias. For models in which a temperature parameter was available (GPT-4.1), temperature was explicitly set to 0 to ensure deterministic and reproducible outputs.

Each gait cycle was classified independently in a single API call. Models were evaluated both with and without an additional reference payload containing population-level statistics of the NORMAL gait class (mean, standard deviation, 5th and 95th percentiles for each feature and time point). Consequently, in contrast to the OCSVM approach, the LLM with reference grounding did not receive raw sample data but only aggregated population statistics derived from the NORMAL class. To prevent data leakage, when a given subject was evaluated, all NORMAL-class gait cycles from that subject were excluded from the computation of the reference statistics. The reference payload was otherwise fixed and identical across calls.

Kinematic features were transformed from flat tabular data into a hierarchically structured JSON format, indexed by biomechanical feature, anatomical side, and temporal gait cycle percentage (e.g., hip_flexion → left → t50). This labeling scheme provides explicit semantic context for each numerical value, which has been shown to improve LLM interpretation of quantitative inputs compared to unlabeled numerical sequences (Gruver et al., 2023; Sui et al., 2023). Furthermore, values were rounded to two decimal places to reduce token usage.

### 2.6.2. Prompt Engineering

Prompt engineering was performed in collaboration with three experts in biomechanical research, each with more than five years of experience in biomechanics and prior experience working with large language models. The experts developed an initial prompt (see Appendix). The prompt design was guided by established best practices in prompt engineering described in the scientific literature, particularly emphasizing structured instructions, explicit task specification, role assignment, constrained output formats, and clearly defined decision rules, which have been shown to improve model reliability, reduce hallucinations, and enhance task-specific performance (Brown et al., 2020; Reynolds & McDonell, 2021; White et al., 2023). The LLM was instructed to assume the role of a clinical gait biomechanics expert and to perform single-trial classification based exclusively on the provided time-normalized kinematic input and, when applicable, population-level reference statistics. The model was required to select exactly one class from a predefined set and to return its output in a strictly defined JSON schema containing: (i) the predicted class label, (ii) a qualitative confidence estimate (high, medium, or low), and (iii) a brief biomechanical justification. Responses not conforming to the predefined JSON schema were automatically resubmitted.

The prompt specified whether deviations were unilateral or bilateral but did not indicate which limb should be inspected. Although unilateral deviations in the dataset consistently affected the right side, this information was intentionally omitted to better reflect realistic clinical scenarios in which the affected side may not be known a priori.

The prompt was piloted using a subset of gait cycles that had been excluded from the final dataset. These pilot samples originated from three randomly selected subjects whose recordings contained a surplus of gait cycles. Results from this pilot phase indicated that the initial class descriptions were insufficiently discriminative. Consequently, the prompt was refined by adding two clarifying elements (highlighted in the Appendix). The finalized prompt was fixed prior to all classification experiments and remained unchanged throughout the study.

## 2.7. Evaluation Metrics and Calculations

All evaluation metrics were computed from aggregated out-of-fold predictions obtained across the LOSO iterations. The MCC was used as the primary performance metric, as it provides a balanced evaluation of classification quality by incorporating all confusion matrix entries and is less susceptible to class imbalance compared to simpler metrics (Chicco et al., 2021). In addition, the macro-averaged F1-score was reported to quantify predictive performance independently of class frequency. This was particularly relevant for the separate analyses stratified by confidence rating, where class distributions could differ substantially.

All analyses, metric calculations, and visualizations were performed in Python using *scikit-learn* (Buitinck et al., 2013), *SciPy* (Virtanen et al., 2020), *seaborn* (Waskom, 2021), and the *OpenAI API* (OpenAI, 2025).

# 3. Results

## 3.1. Reference ML Classifiers

Using the supervised KNN classifier on the multiclass classification task led to an F1 score of 0.90 and MCC of 0.88. When the predictions were projected into the same binary decision space (NORMAL vs. NOT-NORMAL), performance corresponded to an F1 score of 0.86 and an MCC of 0.74. In comparison, the OCSVM classifier showed lower performance in differentiation of NORMAL and NOT-NORMAL (F1 = 0.80, MCC = 0.60). The respective confusion matrices are presented in Figure 2.

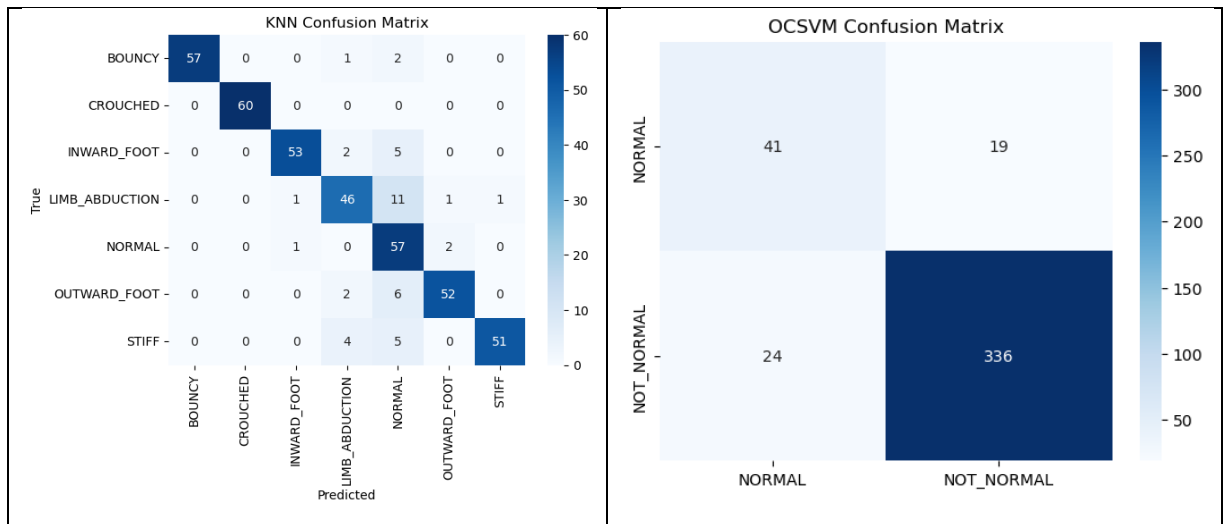

**Figure 2.** Confusion matrices for the supervised KNN multiclass classifier (left) and the class-independent classifier via OCSVM (right).

## 3.2. LLM Classifier

The results for the LLM models, evaluated with and without reference data grounding for the NORMAL gait class, are summarized in Table 2. Across all tested models, grounding consistently and substantially improves classification performance. The highest performance is achieved by GPT-5, followed closely by o4-mini, which shows only a minor reduction in performance. The corresponding confusion matrices are shown in Figure 3.

In comparison to the multiclass KNN model, even the best-performing LLM demonstrates notably lower overall performance.

Table 2. Overall performance of the LLM approaches and results stratified by confidence rating. Confidence-stratified results refer to multiclass classification performance. The highest performance values are highlighted in bold. *Note:* Performance metrics for lower confidence categories are not reported due to insufficient sample sizes. F1 = macro-averaged F1-score; MCC = Matthews correlation coefficient.

| | | | Overall Multi | Overall Binary | High Confidence | Medium Confidence | Low Confidence |
|---|---|---|---|---|---|---|---|
| **with reference data** | o4-mini | % samples | | | 80.24% | 19.52% | 0.24% |
| | | F1 | **0.73** | 0.81 | 0.80 | 0.35 | |
| | | MCC | 0.69 | 0.61 | 0.78 | 0.33 | |
| | GPT-5-mini | % samples | | | 74.29% | 25.71% | 0.00% |
| | | F1 | 0.66 | 0.72 | 0.74 | 0.28 | |
| | | MCC | 0.63 | 0.52 | 0.73 | 0.30 | |
| | GPT-4.1 | % samples | | | 93.79% | 6.21% | 0.00% |
| | | F1 | 0.53 | 0.53 | 0.56 | 0.02 | |
| | | MCC | 0.50 | 0.50 | 0.54 | 0.00 | |
| | GPT-5 | % samples | | | 76.90% | 22.62% | 0.48% |
| | | F1 | **0.73** | **0.83** | **0.89** | 0.28 | |
| | | MCC | **0.70** | **0.68** | **0.83** | 0.27 | |
| **without reference data** | o4-mini | % samples | | | 79.52% | 19.05% | 1.43% |
| | | F1 | 0.57 | 0.57 | 0.64 | 0.19 | |
| | | MCC | 0.50 | 0.50 | 0.59 | 0.07 | |
| | GPT-5-mini | % samples | | | 73.33% | 26.67% | 0.00% |
| | | F1 | 0.37 | 0.37 | 0.41 | 0.30 | |
| | | MCC | 0.39 | 0.39 | 0.44 | 0.23 | |
| | GPT-4.1 | % samples | | | 84.76% | 15.24% | 0.00% |
| | | F1 | 0.36 | 0.60 | 0.34 | 0.09 | |
| | | MCC | 0.40 | 0.22 | 0.43 | 0.00 | |
| | GPT-5 | % samples | | | 72.86% | 26.90% | 0.24% |
| | | F1 | 0.64 | 0.72 | 0.71 | 0.31 | |
| | | MCC | 0.59 | 0.48 | 0.68 | 0.30 | |

An analysis of the LLM's self-rated confidence indicates that for all applied models most predictions are assigned a high confidence level, whereas only a small proportion of samples receive medium or low confidence ratings (Table 2). Performance varies systematically across these confidence categories: predictions assigned high confidence exhibit higher F1 and MCC values, whereas those associated with medium confidence show markedly reduced performance. Low-confidence predictions (when present, as some models do not self-rate with low confidence) are largely non-informative.

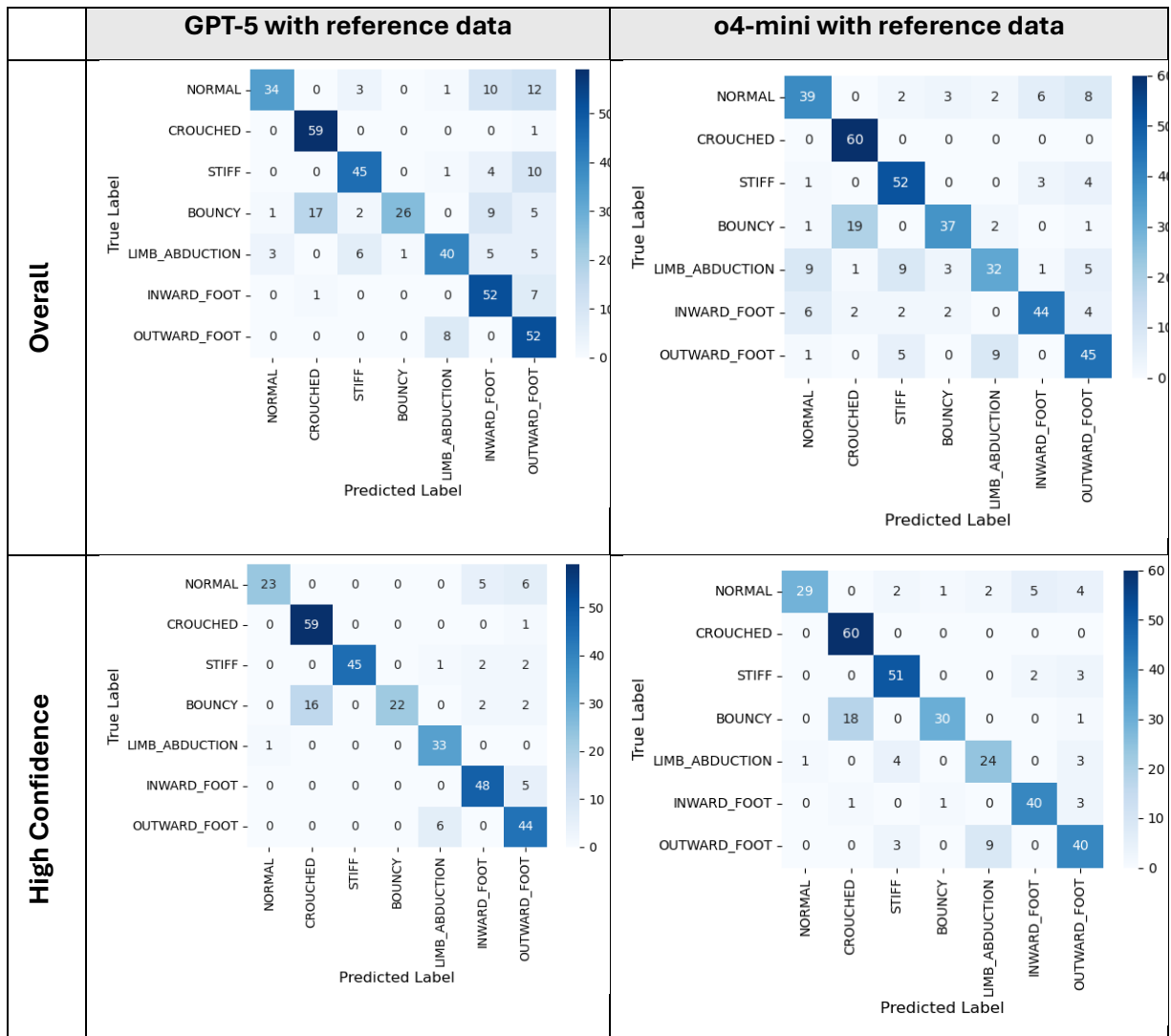

**Figure 3.** Comparison of the two best-performing LLM classifiers with reference grounding. Top: results for all predictions. Bottom: results restricted to high-confidence predictions.

## 3.3. Explorative Analysis of Misclassification Patterns

An explorative analysis of the misclassifications was performed for the best-performing model – GPT-5 with reference grounding. The most frequent misclassification types and representative excerpts from the justifications are summarized in Table 3.

**Table 3.** Recurrent class-wise misclassification patterns and representative justification excerpts from GPT-5 with reference grounding. Quotes were selected from several different samples per pattern to illustrate recurring justification behavior.

| Target → Predicted | Representative excerpts from LLM justifications |
|---|---|
| BOUNCY → CROUCHED | "both knees remain excessively flexed across the cycle, failing to re-extend toward normal during stance"; "persistently elevated sagittal flexion without re-extension"; "hips are likewise overly flexed in stance"; "increased dorsiflexion accompanies sustained knee flexion" |
| NORMAL → OUTWARD_FOOT | "persistent external rotation across most of the gait cycle"; "exceeding normative p95 values"; "external transverse-plane deviation present even during phases where internal rotation is expected" |
| NORMAL → INWARD_FOOT | "persistent internal transverse-plane rotation"; "rotation outside expected normative ranges despite otherwise typical kinematics"; "no dominant frontal-plane deviation present" |
| STIFF → OUTWARD_FOOT | "hip rotation persistently externally rotated across stance"; "transverse-plane deviation dominates despite reduced sagittal excursion"; "external rotation remains evident during mid-stance when normals rotate internally" |
| BOUNCY → INWARD_FOOT | "persistent internal hip rotation beyond normal p5 values"; "transverse-plane deviation more salient than sagittal flexion pattern"; "rotation abnormal across mid-stance to swing" |
| OUTWARD_FOOT → LIMB_ABDUCTION | "marked swing-phase hip abduction"; "excessive unilateral abduction during swing dominates interpretation"; "transverse-plane features less emphasized than frontal-plane deviation" |

The most frequent misclassification was BOUNCY gait predicted as CROUCHED. In these cases, the justifications consistently described sustained bilateral hip and knee flexion across the gait cycle and emphasized the absence of re-extension during stance. These descriptions were linked to wording such as "persistently flexed" and "supports a crouched posture."

For samples labelled NORMAL but predicted as OUTWARD_FOOT or INWARD_FOOT, the justifications focused on persistent external or internal transverse-plane rotation across the gait cycle. In several cases, the model explicitly referred to rotation being used as a proxy for foot progression and stated that this pattern "best fits" an inward- or outward-foot classification, while explicitly excluding crouched, stiff, or bouncy patterns based on the sagittal-plane profiles.

Misclassifications of STIFF as OUTWARD_FOOT were characterized by descriptions of persistent external hip rotation, accompanied by explicit statements that crouched or bouncy sagittal patterns were not evident. The outward-foot classification was justified primarily by the presence of transverse-plane rotation.

For BOUNCY samples misclassified as INWARD_FOOT, the justifications again emphasized persistent internal rotation and explicitly described hip rotation as a proxy for foot progression, leading to the assignment of an inward-foot label.

Finally, OUTWARD_FOOT samples misclassified as LIMB_ABDUCTION were justified by repeated references to pronounced unilateral hip abduction during swing phase, which was described as dominating the interpretation.

## 4. Discussion

We evaluated OpenAI's contemporary general-purpose LLMs on continuous biomechanical gait waveforms presented as textual numeric tokens and compared zero-shot LLM behaviour to both supervised multiclass classifiers and a class-independent OCSVM. Using systematically induced gait variations as a high–internal-validity testbed, the supervised KNN model yielded the highest multiclass classification performance. Because the model was trained with complete class label information, its performance may be considered an approximate upper-bound reference, representing the best achievable accuracy under fully supervised conditions in this setting. Among the LLM approaches, GPT-5 with reference grounding achieved the highest accuracy and, when its outputs were aligned to a binary decision space (NORMAL vs. NOT-NORMAL), it slightly outperformed the one-class anomaly detection method on this binary classification task (RQ1).

The marked performance gap between the supervised multiclass classifier and the LLMs highlights how strongly this task aligns with the inductive biases (i.e., the structural assumptions a model makes about how data is organized) of models that operate directly in continuous numerical feature spaces. KNN operates on numerical representations in which the geometric and temporal structure of joint kinematics remains intact, allowing distance-based comparisons to exploit the spatial and temporal relationships present across the gait cycle. This close alignment between data representation and model operation is consistent with prior work demonstrating how architectural inductive biases determine task suitability across model classes (Tay et al., 2023).

In contrast, LLMs are architecturally and statistically optimized to process symbolic sequences that follow linguistic structure (Spathis & Kawsar, 2024). When continuous kinematic waveforms are represented as sequences of numerical tokens, the spatial and temporal structure of the signal is no longer preserved in a form that matches these inductive biases. Instead, the models process the data as symbolic elements in a sequence rather than as geometrically and temporally organized measurements. While prior work has shown that LLMs can act as zero-shot time-series forecasters by recognizing repetitive or patterned sequences (Gruver et al., 2023), such capabilities rely on detecting token-level statistical regularities rather than modeling the underlying geometric or dynamical structure of the signal.

These findings therefore underscore that, for tasks requiring fine-grained spatio-temporal discrimination of biomechanical signals, comparatively simple models that operate natively in numerical feature spaces may remain better suited than highly complex language models whose strengths lie in processing symbolic, linguistically structured data.

Examination of LLM predictions and accompanying explanations revealed a consistent salience bias (Clark et al., 2026) (i.e., a tendency to focus on the most numerically or visually prominent features while underweighting subtler but diagnostically relevant patterns). Across misclassifications, model outputs were systematically dominated by amplitude-based features, while subtler phase-dependent relationships and diagnostically relevant absences of expected motion were underweighted. A recurring pattern was observed in the confusion between BOUNCY and CROUCHED gait: interpretations were frequently driven by exaggerated knee flexion, despite the presence of clear re-extension phases that are diagnostically critical for correct discrimination. This pattern indicates that the LLM relied more heavily on numerically prominent deviations than on the temporal coordination of joint waveforms.

In general, the interpretation of LLM explanations for classification decisions requires caution, as recent analyses have shown that such self-explanations can be fluent and persuasive without being fully faithful to the model's internal decision process (Madsen et al., 2024). Consequently, the plausibility of an explanation does not guarantee faithfulness, and these explanations may inadvertently mislead clinical users if presented without appropriate caveats.

The best-performing LLM configuration (GPT-5 with reference grounding and decision-space alignment to a one-class formulation) slightly exceeded the OCSVM's performance in the task of distinguishing NORMAL from NOT-NORMAL observations. Although prior studies have shown that OCSVMs can approach the performance of supervised classifiers in gait analysis (Dindorf et al., 2021), the OCSVM in the present study underperformed relative to the supervised benchmark.

This gap can potentially be interpreted through the lens of how one-class methods represent "normality." OCSVMs estimate a compact support of the training distribution in feature space and flag deviations that fall outside this learned boundary (Schölkopf et al., 2001). Their effectiveness therefore depends critically on how well the chosen features separate NORMAL from NOT-NORMAL gait patterns. Subtle pathological deviations may remain numerically similar to normal patterns in this representation, in which case boundary-based detectors can struggle.

By contrast, the LLM did not operate by learning a geometric boundary around normal samples. Instead, it had access to additional sources of information, including prior knowledge encoded during pretraining and the provided linguistic class descriptions, whereas the OCSVM relied exclusively on the geometry of the observed feature space. From this perspective, the LLM's performance should be interpreted cautiously: despite benefiting from substantially richer informational context, it did not clearly outperform the comparatively simple one-class method. While LLMs are not suitable substitutes for supervised multiclass classifiers in fine-grained gait classification, their performance in the anomaly-style setting should be interpreted with caution. Despite access to substantially richer informational context than the OCSVM, the LLM achieved only a modest advantage, indicating that such models do not automatically translate their pretrained knowledge into superior performance in this domain.

Although GPT-5 achieved the highest overall performance, the smaller o4-mini model performed comparably, while GPT-5-mini and GPT-4.1 showed substantially lower performance despite being newer or larger. This pattern indicates that, for the present task, performance did not scale monotonically with parameter count or model recency; however, as only OpenAI models were evaluated, this finding should be interpreted within this specific model family and task context.

The relatively strong performance of o4-mini suggests that factors beyond scale — including differences in training emphasis or internal representation strategies — may influence how effectively LLMs handle structured numerical inputs. This observation aligns with research showing that tokenization and intermediate representation choices, independent of model size, can materially affect downstream performance in LLMs (Ali et al., 2023). Taken together, these findings suggest that targeted improvements in how continuous biomechanical signals are represented for LLM processing may be a more promising avenue for enhancing performance than increasing model scale alone.

The explicit provision of reference grounding was associated with a large and consistent improvement in LLM accuracy (RQ2). This finding is consistent with prior work suggesting that general-purpose LLMs benefit from explicit contextual anchors when applied to specialized biomedical domains (K. He et al., 2025). It also aligns with research on retrieval-augmented generation, which shows that supplying task-relevant external information at inference time can improve domain performance. In the present study, this grounding was achieved through static prompt design rather than an automated retrieval pipeline (Liu et al., 2025).

When provided with a structured reference frame, the LLMs may have been better able to relate the numerical inputs to meaningful contextual patterns, potentially shifting the task from isolated numerical classification toward a more comparative, context-guided form of reasoning. Accordingly, in biomechanical applications, grounding may offer a practical way to supply reference distributions and thresholds that general-purpose LLMs might not otherwise utilize effectively when interpreting structured numerical kinematic gait data. However, in many real-world biomedical settings, such normative reference values may not be available to include in prompts. In such cases, users may need to rely primarily on the model's internal representation of the task, which may introduce uncertainty regarding performance in practical applications.

Although previous research has shown limited calibration between self-reported confidence and true accuracy across a range of LLMs (Naderi et al., 2026; Omar et al., 2025), the present results demonstrate that performance varies systematically across confidence categories (RQ3). For all tested LLM configurations, predictions labeled as high confidence were associated with markedly higher empirical performance than those assigned medium or low confidence. Notably, 76.90% of all samples were classified with high model-reported confidence using the best-performing LLM configuration, and this subset exhibited comparatively strong empirical performance (MCC = 0.83). However, this level of performance must be interpreted in the context of the simplified and highly controlled task design of the present study, which is substantially less complex than real-world gait assessment scenarios. The fact that misclassifications still occurred within this high-confidence group under these favorable conditions indicates that model-reported confidence constitutes an informative but imperfect proxy for correctness rather than a guarantee of validity, and its practical utility in more complex settings may therefore be limited.

At the same time, the persuasive nature of LLM-generated explanations warrants additional caution. Prior work in clinical reasoning has shown that models frequently express high certainty even when incorrect, and that fluent justifications can create an unwarranted sense of correctness despite incomplete or flawed reasoning (Omar et al., 2025). In this context, self-reported high confidence, combined with coherent explanations, may unintentionally reinforce misplaced trust if interpreted without critical oversight.

The present findings must be interpreted within several important limitations. The reliance on simulated, mechanically defined deviations prioritized internal validity but did not capture the complexity of real clinical gait, where overlapping pathologies and compensatory strategies can obscure primary deviations (Schmidt et al., 2019). Prompt design represents an additional methodological trade-off. Detailed biomechanical class descriptions and explicit guidance were included due to overlapping gait characteristics but may have implicitly directed the model toward specific features and reduced the effective difficulty of the task. In addition, prompt development used gait cycles from three subjects with surplus recordings. Although these samples were excluded from the final evaluation set, within-subject gait cycles are biomechanically correlated. This may have led to a slight overestimation of the LLM's classification performance due to indirect familiarity with subject-specific motion patterns during prompt refinement. The reported confidence estimate was explicitly elicited within the prompt and reflects the model's self-reported, qualitative confidence. No token-level log probabilities or calibrated probability scores were extracted or used for confidence estimation. Thus, the confidence ratings should be interpreted as subjective model assessments rather than statistically calibrated uncertainty measures. Furthermore, the inherent prompt dependence of LLMs raises reproducibility concerns, as variations in prompt design and structure can materially shift classification behavior, a phenomenon observed in evaluations of LLM performance on radiology report classification tasks (Sorin et al., 2025).

The observed performance gap between general-purpose LLMs and supervised classifiers (RQ1) can be interpreted as reflecting a representational mismatch. Recent frameworks have begun to treat biomechanical and human motion signals as a modal complement to language by using hybrid architectures that align continuous motion embeddings with language model representations (i.e., models that jointly encode motion and text for understanding and generation (Yang et al., 2025; Zhu et al., 2025). Such task-optimized multimodal models may represent a more sustainable path forward than relying solely on general-purpose text-based LLMs.

Finally, translating these findings into real-world clinical settings requires increased attention to human-factors research. It is essential to evaluate clinician trust, the interpretability of LLM-generated biomechanical reasoning within diagnostic workflows, and whether the observed "reasoning" translates into improved clinical decision-making in real-world settings.

## 5. Conclusion

This study shows that, under the specific prompt design and experimental setup used here, general-purpose LLMs did not achieve the performance of supervised multiclass classifiers for precise gait classification when continuous kinematic waveforms were presented as textual numeric tokens. Although their performance approached that of a one-class anomaly detection method after alignment to a binary decision space, this comparison must be interpreted cautiously given the additional contextual information available to the LLM, the simplicity of the present task, and the known sensitivity of LLM performance to prompt formulation and category definitions. Furthermore, while LLMs generated fluent explanations alongside their predictions, these explanations cannot be assumed to faithfully reflect the model's internal decision process and may inadvertently encourage overtrust if interpreted uncritically. Overall, the findings indicate that general-purpose LLMs, in their current form, should not be considered diagnostic tools for gait analysis but rather exploratory systems whose outputs require careful human oversight and critical interpretation.

# Data availability

The datasets used during the current study are available from the corresponding author on reasonable request.

# Appendix

> You are a clinical gait biomechanics expert specializing in kinematic pattern recognition.
>
> Your task is to classify a single gait trial.
>
> ### You are given:
> 1) TRIAL DATA (% Gait Cycle): Time-normalized joint kinematics for one gait trial (0–100%), starting at initial contact of the right limb.
> 2) REFERENCE STATS: Population statistics for NORMAL gait class (mean, SD, ranges).
>
> ### Your role:
> - Compare TRIAL DATA against REFERENCE STATS.
> - Identify the dominant biomechanical deviation pattern.
> - Select the single BEST-FITTING class.
> - Return ONLY valid JSON. No markdown. No extra text.
>
> ### Exact output schema:
> {{
>   "class": "<one allowed class label; do NOT include descriptions or additional text>",
>   "confidence": "<high | medium | low>",
>   "justification": "<3–6 concise sentences explaining which kinematic features and gait-cycle phases support the decision>"
> }}
>
> ### Allowed classes (use strictly):
> 1. NORMAL: Habitual healthy gait at 4 km/h.
> 2. BOUNCY: Bilateral gait with exaggerated flexion followed by a clear, rapid re-extension within the same gait cycle. The joints return toward normal extension during stance, producing a pronounced "bouncy" pattern.
> 3. CROUCHED: Bilateral gait characterized by persistent hyper-flexion across stance and swing. The joints fail to re-extend toward normal, remaining continuously flexed throughout the gait cycle.
> 4. STIFF: Bilateral stiffness of the knee during the gait cycle in sagittal plane, not reaching normal flexion maximum.

> 5. LIMB_ABDUCTION: Unilateral excessive limb abduction occurring exclusively during the swing phase.
> 6. OUTWARD_FOOT: Persistent unilateral external rotation of the foot in transversal plane across most or all phases of the gait cycle, WITHOUT a dominant frontal-plane limb abduction pattern during swing.
> 7. INWARD_FOOT: Persistent unilateral internal rotation of the foot in transversal plane across most or all phases of the gait cycle.
>
> ### Clinical decision rules:
> - Classify as NORMAL when deviations are minor, isolated, or inconsistent and no other dominant class pattern is present.
> - If a clear unilateral pattern is present, prioritize it over any concurrent bilateral pattern.
> - Confidence reflects strength of evidence: high (strong, consistent), medium (moderate), low (weak or mixed).
>
> ### Kinematic feature interpretation rules:
> - Only the kinematic variables explicitly provided in TRIAL DATA are available for interpretation; assume no additional degrees of freedom exist.
> - When a class refers to a motion not directly represented by a provided variable, use the most biomechanically appropriate proxy among the available features.
>
> ### Coordinate system:
> - Hip rotation: Negative values indicate internal rotation, positive values indicate external rotation, independent of limb side.
> - Hip adduction/abduction: Negative values indicate adduction, positive values indicate abduction, independent of limb side.
>
> ### Data inputs:
> - REFERENCE STATS: {reference_text}
> - TRIAL DATA (% Gait Cycle): {feature_text}

**Figure**. Final prompt used. Yellow highlighted are text elements added by the experts after pilot testing. Blue highlighted are text elements included for analyzing research question 2 including the references data of the normal gait.

# References

Ali, M., Fromm, M., Thellmann, K., Rutmann, R., Lübbering, M., Leveling, J., Klug, K., Ebert, J., Doll, N., Buschhoff, J. S., Jain, C., Weber, A. A., Jurkschat, L., Abdelwahab, H., John, C., Suarez, P. O., Ostendorff, M., Weinbach, S., Sifa, R., … Flores-Herr, N. (2023). *Tokenizer Choice For LLM Training: Negligible or Crucial?* (Version 4). arXiv. https://doi.org/10.48550/ARXIV.2310.08754

Balaji, Brindha, & Balakrishnan. (2020). Supervised machine learning based gait classification system for early detection and stage classification of Parkinson's disease. *Applied Soft Computing*, *94*, 106494. https://doi.org/10.1016/j.asoc.2020.106494

Brown, T. B., Mann, B., Ryder, N., Subbiah, M., Kaplan, J., Dhariwal, P., Neelakantan, A., Shyam, P., Sastry, G., Askell, A., Agarwal, S., Herbert-Voss, A., Krueger, G., Henighan, T., Child, R., Ramesh, A., Ziegler, D. M., Wu, J., Winter, C., … Amodei, D. (2020). *Language Models are Few-Shot Learners* (Version 4). arXiv. https://doi.org/10.48550/ARXIV.2005.14165

Buitinck, L., Louppe, G., Blondel, M., Pedregosa, F., Mueller, A., Grisel, O., Niculae, V., Prettenhofer, P., Gramfort, A., Grobler, J., Layton, R., Vanderplas, J., Joly, A., Holt, B., & Varoquaux, G. (2013). *API design for machine learning software: Experiences from the scikit-learn project*. https://doi.org/10.48550/ARXIV.1309.0238

Cash, T. N., Oppenheimer, D. M., Christie, S., & Devgan, M. (2025). Quantifying uncert-AI-nty: Testing the accuracy of LLMs' confidence judgments. *Memory & Cognition*. https://doi.org/10.3758/s13421-025-01755-4

Chicco, D., Tötsch, N., & Jurman, G. (2021). The Matthews correlation coefficient (MCC) is more reliable than balanced accuracy, bookmaker informedness, and markedness in two-class confusion matrix evaluation. *BioData Mining*, *14*(1), 13. https://doi.org/10.1186/s13040-021-00244-z

Clark, B., Oliveira, M., Wilming, R., & Haufe, S. (2026). *Feature salience—Not task-informativeness—Drives machine learning model explanations* (Version 1). arXiv. https://doi.org/10.48550/ARXIV.2602.09238

Delp, S. L., Anderson, F. C., Arnold, A. S., Loan, P., Habib, A., John, C. T., Guendelman, E., & Thelen, D. G. (2007). OpenSim: Open-Source Software to Create and Analyze Dynamic Simulations of Movement. *IEEE Transactions on Biomedical Engineering*, *54*(11), 1940–1950. https://doi.org/10.1109/TBME.2007.901024

Delp, S. L., Loan, J. P., Hoy, M. G., Zajac, F. E., Topp, E. L., & Rosen, J. M. (1990). An interactive graphics-based model of the lower extremity to study orthopaedic surgical procedures. *IEEE Transactions on Biomedical Engineering*, *37*(8), 757–767. https://doi.org/10.1109/10.102791

Dindorf, C., Horst, F., Slijepčević, D., Dumphart, B., Dully, J., Zeppelzauer, M., Horsak, B., & Fröhlich, M. (2025). Machine Learning in Biomechanics: Key Applications and Limitations in Walking, Running and Sports Movements. In M. J. Blondin, I. Fister, & P. M. Pardalos (Eds.), *Artificial Intelligence, Optimization, and Data Sciences in Sports* (Vol. 218, pp. 91–148). Springer Nature Switzerland. https://doi.org/10.1007/978-3-031-76047-1_4

Dindorf, C., Konradi, J., Wolf, C., Taetz, B., Bleser, G., Huthwelker, J., Werthmann, F., Bartaguiz, E., Kniepert, J., Drees, P., Betz, U., & Fröhlich, M. (2021). Classification and Automated Interpretation of Spinal Posture Data Using a Pathology-Independent Classifier and Explainable Artificial Intelligence (XAI). *Sensors*, *21*(18), 6323. https://doi.org/10.3390/s21186323

Dindorf, C., Teufl, W., Taetz, B., Bleser, G., & Fröhlich, M. (2020). Interpretability of Input Representations for Gait Classification in Patients after Total Hip Arthroplasty. *Sensors*, *20*(16), 4385. https://doi.org/10.3390/s20164385

Eltanani, S., Scheper, T. O., & Dawes, H. (2021). K - Nearest Neighbor Algorithm: Proposed Solution for Human Gait Data Classification. *2021 IEEE Symposium on Computers and Communications (ISCC)*, 1–5. https://doi.org/10.1109/ISCC53001.2021.9631454

Gruver, N., Finzi, M., Qiu, S., & Wilson, A. G. (2023). *Large Language Models Are Zero-Shot Time Series Forecasters* (Version 3). arXiv. https://doi.org/10.48550/ARXIV.2310.07820

Halilaj, E., Rajagopal, A., Fiterau, M., Hicks, J. L., Hastie, T. J., & Delp, S. L. (2018). Machine learning in human movement biomechanics: Best practices, common pitfalls, and new opportunities. *Journal of Biomechanics*, *81*, 1–11. https://doi.org/10.1016/j.jbiomech.2018.09.009

He, K., Mao, R., Lin, Q., Ruan, Y., Lan, X., Feng, M., & Cambria, E. (2025). A survey of large language models for healthcare: From data, technology, and applications to accountability and ethics. *Information Fusion*, *118*, 102963. https://doi.org/10.1016/j.inffus.2025.102963

He, W., Jiang, Z., Xiao, T., Xu, Z., & Li, Y. (2023). *A Survey on Uncertainty Quantification Methods for Deep Learning* (Version 7). arXiv. https://doi.org/10.48550/ARXIV.2302.13425

Horsak, B., Simonlehner, M., Quehenberger, V., Dumphart, B., Slijepčević, D., & Kranzl, A. (2025). A gait lab in your pocket? Accuracy and reliability of monocular smartphone-based markerless 3D gait analysis in pathological gait. *Gait & Posture*, *121*, 91–92. https://doi.org/10.1016/j.gaitpost.2025.07.102

Horst, F., Lapuschkin, S., Samek, W., Müller, K.-R., & Schöllhorn, W. I. (2019). Explaining the unique nature of individual gait patterns with deep learning. *Scientific Reports*, *9*(1), 2391. https://doi.org/10.1038/s41598-019-38748-8

Keilhauer, R., Lorenz, M., Dindorf, C., Ernst, S., Wang, C.-Y., Messer, P., & Stricker, D. (2025). Exploring Large Language Models for Automated Gait Analysis. *2025 Second International Conference on Artificial Intelligence for Medicine, Health and Care (AIxMHC)*, 136–142. https://doi.org/10.1109/AIxMHC65380.2025.00031

Kirtley, C. (2006). *Clinical gait analysis: Theory and practice*. Elsevier.

Kwon, S. B., Han, H.-S., Lee, M. C., Kim, H. C., Ku, Y., & Ro, D. H. (2020). Machine Learning-Based Automatic Classification of Knee Osteoarthritis Severity Using Gait Data and Radiographic Images. *IEEE Access*, *8*, 120597–120603. https://doi.org/10.1109/ACCESS.2020.3006335

Liu, S., McCoy, A. B., & Wright, A. (2025). Improving large language model applications in biomedicine with retrieval-augmented generation: A systematic review, meta-analysis, and clinical development guidelines. *Journal of the American Medical Informatics Association*, *32*(4), 605–615. https://doi.org/10.1093/jamia/ocaf008

Madsen, A., Chandar, S., & Reddy, S. (2024). Are self-explanations from Large Language Models faithful? *Findings of the Association for Computational Linguistics ACL 2024*, 295–337. https://doi.org/10.18653/v1/2024.findings-acl.19

Mezghani, N., Husse, S., Boivin, K., Turcot, K., Aissaoui, R., Hagemeister, N., & De Guise, J. A. (2008). Automatic Classification of Asymptomatic and Osteoarthritis Knee Gait Patterns Using Kinematic Data Features and the Nearest Neighbor Classifier. *IEEE Transactions on Biomedical Engineering*, *55*(3), 1230–1232. https://doi.org/10.1109/TBME.2007.905388

Naderi, N., Safavi-Naini, S. A. A., Savage, T., Khalafi, M. A., Lewis, P. R., Atf, Z., Nadkarni, G., & Soroush, A. (2026). Across generations, sizes, and types, large language models poorly report self-confidence in gastroenterology clinical reasoning tasks. *Npj Gut and Liver*, *3*(1), 6. https://doi.org/10.1038/s44355-026-00053-3

Omar, M., Agbareia, R., Glicksberg, B. S., Nadkarni, G. N., & Klang, E. (2025). Benchmarking the Confidence of Large Language Models in Answering Clinical Questions: Cross-Sectional Evaluation Study. *JMIR Medical Informatics*, *13*, e66917–e66917. https://doi.org/10.2196/66917

OpenAI. (2025). *OpenAI API* [Computer software]. https://platform.openai.com

Peimankar, A., Winther, T. S., Ebrahimi, A., & Wiil, U. K. (2023). A Machine Learning Approach for Walking Classification in Elderly People with Gait Disorders. *Sensors*, *23*(2), 679. https://doi.org/10.3390/s23020679

Peterson, L. (2009). K-nearest neighbor. *Scholarpedia*, *4*(2), 1883. https://doi.org/10.4249/scholarpedia.1883

Reynolds, L., & McDonell, K. (2021). *Prompt Programming for Large Language Models: Beyond the Few-Shot Paradigm* (Version 1). arXiv. https://doi.org/10.48550/ARXIV.2102.07350

Schmidt, R. A., Lee, T. D., Winstein, C. J., Wulf, G., & Zelaznik, H. N. (2019). *Motor control and learning: A behavioral emphasis* (Sixth edition). Human Kinetics.

Schölkopf, B., Platt, J. C., Shawe-Taylor, J., Smola, A. J., & Williamson, R. C. (2001). Estimating the Support of a High-Dimensional Distribution. *Neural Computation*, *13*(7), 1443–1471. https://doi.org/10.1162/089976601750264965

Singh, C., Morris, J. X., Aneja, J., Rush, A. M., & Gao, J. (2023). *Explaining Patterns in Data with Language Models via Interpretable Autoprompting* (arXiv:2210.01848). arXiv. https://doi.org/10.48550/arXiv.2210.01848

Sorin, V., Collins, J. D., Bratt, A. K., Kusmirek, J. E., Mugu, V. K., Kline, T. L., Butler, C. L., Wood, N. G., Cook, C. J., & Korfiatis, P. (2025). Evaluating prompt and data perturbation sensitivity in large language models for radiology reports classification. *JAMIA Open*, *8*(4), ooaf073. https://doi.org/10.1093/jamiaopen/ooaf073

Spathis, D., & Kawsar, F. (2024). The first step is the hardest: Pitfalls of representing and tokenizing temporal data for large language models. *Journal of the American Medical Informatics Association: JAMIA*, *31*(9), 2151–2158. https://doi.org/10.1093/jamia/ocae090

Stetter, B. J., Dully, J., Stief, F., Holder, J., Steingrebe, H., Zaucke, F., Sell, S., Van Drongelen, S., & Stein, T. (2025). Explainable machine learning for orthopedic decision-making: Predicting functional outcomes of total hip replacement from gait biomechanics. *Arthritis Research & Therapy*, *27*(1), 229. https://doi.org/10.1186/s13075-025-03709-2

Stetter, B. J., & Stein, T. (2024). Machine Learning in Biomechanics: Enhancing Human Movement Analysis. In C. Dindorf, E. Bartaguiz, F. Gassmann, & M. Fröhlich (Eds.), *Artificial Intelligence in Sports, Movement, and Health* (pp. 139–160). Springer Nature Switzerland. https://doi.org/10.1007/978-3-031-67256-9_9

Sui, Y., Zhou, M., Zhou, M., Han, S., & Zhang, D. (2023). *Table Meets LLM: Can Large Language Models Understand Structured Table Data? A Benchmark and Empirical Study* (Version 5). arXiv. https://doi.org/10.48550/ARXIV.2305.13062

Szabó, A., & Laein, G. D. (2025). Comparative evaluation of large language models performance in medical education using urinary system histology assessment. *Scientific Reports*, *15*(1), 31933. https://doi.org/10.1038/s41598-025-17571-4

Tay, Y., Dehghani, M., Abnar, S., Chung, H., Fedus, W., Rao, J., Narang, S., Tran, V., Yogatama, D., & Metzler, D. (2023). Scaling Laws vs Model Architectures: How does Inductive Bias Influence Scaling? *Findings of the Association for Computational Linguistics: EMNLP 2023*, 12342–12364. https://doi.org/10.18653/v1/2023.findings-emnlp.825

Teufl, W., Taetz, B., Miezal, M., Dindorf, C., Fröhlich, M., Trinler, U., Hogan, A., & Bleser, G. (2021). Automated detection and explainability of pathological gait patterns using a one-class support vector machine trained on inertial measurement unit based gait data. *Clinical Biomechanics*, *89*, 105452. https://doi.org/10.1016/j.clinbiomech.2021.105452

Virtanen, P., Gommers, R., Oliphant, T. E., Haberland, M., Reddy, T., Cournapeau, D., Burovski, E., Peterson, P., Weckesser, W., Bright, J., Van Der Walt, S. J., Brett, M., Wilson, J., Millman, K. J., Mayorov, N., Nelson, A. R. J., Jones, E., Kern, R., Larson, E., … Vázquez-Baeza, Y. (2020). SciPy 1.0: Fundamental algorithms for scientific computing in Python. *Nature Methods*, *17*(3), 261–272. https://doi.org/10.1038/s41592-019-0686-2

Wang, D., Bobenrieth, C., & Seo, H. (2025). *AGIR: Assessing 3D Gait Impairment with Reasoning based on LLMs* (arXiv:2503.18141). arXiv. https://doi.org/10.48550/arXiv.2503.18141

Wang, S., Liu, Q., Zhu, E., Porikli, F., & Yin, J. (2018). Hyperparameter selection of one-class support vector machine by self-adaptive data shifting. *Pattern Recognition*, *74*, 198–211. https://doi.org/10.1016/j.patcog.2017.09.012

Waskom, M. (2021). seaborn: Statistical data visualization. *Journal of Open Source Software*, *6*(60), 3021. https://doi.org/10.21105/joss.03021

White, J., Fu, Q., Hays, S., Sandborn, M., Olea, C., Gilbert, H., Elnashar, A., Spencer-Smith, J., & Schmidt, D. C. (2023). *A Prompt Pattern Catalog to Enhance Prompt Engineering with ChatGPT* (Version 1). arXiv. https://doi.org/10.48550/ARXIV.2302.11382

Workum, J. D., Volkers, B. W. S., Van De Sande, D., Arora, S., Goeijenbier, M., Gommers, D., & Van Genderen, M. E. (2025). Comparative evaluation and performance of large language models on expert level critical care questions: A benchmark study. *Critical Care*, *29*(1), 72. https://doi.org/10.1186/s13054-025-05302-0

Xiang, L., Gao, Z., Yu, P., Fernandez, J., Gu, Y., Wang, R., & Gutierrez-Farewik, E. M. (2025). Explainable artificial intelligence for gait analysis: Advances, pitfalls, and challenges - a systematic review. *Frontiers in Bioengineering and Biotechnology*, *13*, 1671344. https://doi.org/10.3389/fbioe.2025.1671344

Xiang, L., Wang, A., Gu, Y., Zhao, L., Shim, V., & Fernandez, J. (2022). Recent Machine Learning Progress in Lower Limb Running Biomechanics With Wearable Technology: A Systematic Review. *Frontiers in Neurorobotics*, *16*, 913052. https://doi.org/10.3389/fnbot.2022.913052

Xu, Q., Peng, Y., Nastase, S. A., Chodorow, M., Wu, M., & Li, P. (2025). Large language models without grounding recover non-sensorimotor but not sensorimotor features of human concepts. *Nature Human Behaviour*, *9*(9), 1871–1886. https://doi.org/10.1038/s41562-025-02203-8

Yang, R., Kennedy, A., & Cotton, R. J. (2025). *BiomechGPT: Towards a Biomechanically Fluent Multimodal Foundation Model for Clinically Relevant Motion Tasks* (Version 1). arXiv. https://doi.org/10.48550/ARXIV.2505.18465

Zhu, B., Jiang, B., Wang, S., Tang, S., Chen, T., Luo, L., Zheng, Y., & Chen, X. (2025). *MotionGPT3: Human Motion as a Second Modality* (arXiv:2506.24086). arXiv. https://doi.org/10.48550/arXiv.2506.24086

# Acknowledgement

The authors would like to express their sincere gratitude to Fabian Golz for his invaluable support throughout the data recording and organization.

# Funding

This research was supported by the Central Innovation Program for Small and Medium-Sized Enterprises (Zentrales Innovationsprogram Mittelstand, ZIM) of the German Federal Ministry for Economic Affairs and Climate Action under Grant number 16KN113027.

# Contributions

Carlo Dindorf: Conceptualization, Methodology, Software, Formal analysis, Investigation, Writing – Original Draft, Writing – Review & Editing, Funding acquisition. Jonas Dully: Investigation, Data curation, Resources, Writing – Original Draft, Writing – Review & Editing. Rebecca Keilhauer: Conceptualization, Methodology, Writing – Review & Editing. Michael Lorenz: Conceptualization, Methodology, Writing – Review & Editing. Michael Fröhlich: Supervision, Project administration, Writing – Review & Editing, Funding acquisition.

# Ethics declarations

## Ethics approval and consent to participate

The study was conducted in accordance with the Declaration of Helsinki and approved by the institutional ethics committee (No. 99). All participants were informed about the study procedures and provided written informed consent prior to experimental procedures.

## Consent for publication

Not applicable.

## Competing interests

The authors declare no competing interests.